\documentclass[conference]{IEEEtran}
\IEEEoverridecommandlockouts
\usepackage[T1]{fontenc}
\usepackage{cite}
\usepackage{amsmath,amssymb,amsfonts}
\usepackage{graphicx}
\usepackage{textcomp}
\usepackage{xcolor}
\usepackage{booktabs}
\usepackage{balance}
\usepackage{microtype}
\usepackage{placeins}
\usepackage{url}
\usepackage{hyperref}
\hypersetup{hidelinks}
\graphicspath{{figures/}}

\def\BibTeX{{\rm B\kern-.05em{\sc i\kern-.025em b}\kern-.08em
    T\kern-.1667em\lower.7ex\hbox{E}\kern-.125emX}}

\title{DYAD: A Multimodal Dataset of\\
Co-Located Human Assistance}

\author{
\IEEEauthorblockN{Akhil Ajikumar\IEEEauthorrefmark{1}, Mahya Qorbani\IEEEauthorrefmark{1},
Sakib Reza\IEEEauthorrefmark{2}, Sean Andrist\IEEEauthorrefmark{3},
Mohsen Moghaddam\IEEEauthorrefmark{1}}
\IEEEauthorblockA{\IEEEauthorrefmark{1}Georgia Institute of Technology, Atlanta, GA, USA}
\IEEEauthorblockA{\IEEEauthorrefmark{2}Northeastern University, Boston, MA, USA}
\IEEEauthorblockA{\IEEEauthorrefmark{3}Microsoft Research, Redmond, WA, USA}
\IEEEauthorblockA{\texttt{\{akhil.aji,mqorbani3,mohsen.moghaddam\}@gatech.edu}\\
\texttt{reza.s@northeastern.edu}, \texttt{sandrist@microsoft.com}}
}

\begin{document}
\bstctlcite{IEEEexample:BSTcontrol}

\maketitle

\begin{abstract}
An embodied assistant working beside a person must track task state, recognize help seeking, choose how to intervene, and produce an appropriate response. Existing procedural datasets richly describe individual execution, while interactive datasets capture remote verbal instruction or undifferentiated co-working. They do not jointly link a co-located helper's verbal and physical interventions to performer requests, task state, assistance triggers, and outcomes. We introduce DYAD (DYadic Assistance Dataset), a synchronized multimodal record of human-human assistance during gearbox assembly. Across 20 sessions, one trained helper follows a guidance-first policy while assisting HoloLens~2 wearers. DYAD links 528 task-step intervals and 611 performer requests with 851 valid assistance records spanning verbal and physical help. DYAD's annotations span the assistance process; three reference tasks evaluate selected components rather than an end-to-end system: causal step understanding, pre-onset mode anticipation, and instructor response generation. On 829 eligible mode events, the strongest four-seed RGB mean is $0.548{\pm}0.007$ macro-F1; causal metadata reaches 0.624 and a privileged trigger mapping 0.915, revealing information not recovered from pre-onset RGB. DYAD's contribution is not scale, but a linked interaction structure spanning help seeking, intervention choice, execution, and outcome under egocentric and workspace sensing.
\end{abstract}


\section{Introduction}
\label{sec:introduction}

\begin{figure*}[t]
    \centering
    \includegraphics[width=0.94\textwidth]{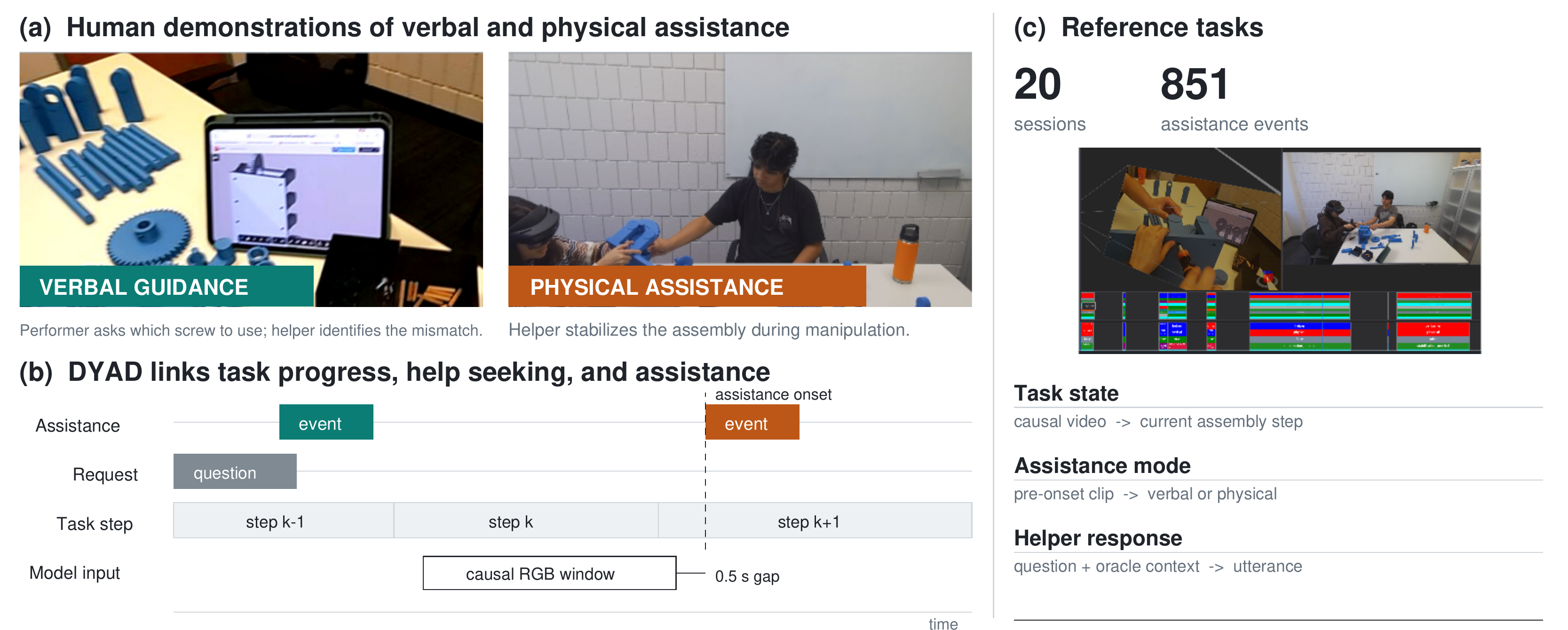}
    \caption{DYAD records co-located performer--helper assembly on a synchronized
    timeline of task steps, requests, and verbal or physical interventions. Three
    reference tasks evaluate selected components, not an end-to-end assistant.}
    \label{fig:teaser}
\end{figure*}

A central decision for a co-located assistive system is whether to speak or act, and when either response is appropriate. Making that decision requires estimating task state and need before selecting and producing a response. Human helpers demonstrate this process by answering questions, waiting for self-correction, handing over unreachable parts, or stabilizing an assembly. Such interactions can supervise robotic and augmented-reality (AR) assistance, although human actions are not automatically safe or robot-executable.

Mode matters: speech cannot hand over an unreachable part, whereas unnecessary physical correction can undermine autonomy. Responses also depend on task state, request referents, and interaction history, so assistance data must link task, request, intervention, and outcome.

Existing datasets cover pieces of this loop. Procedural datasets describe individual actions, steps, and errors~\cite{sener2022assembly101,ben2021ikea,schoonbeek2024industreal,grauman2022ego4d}; Ego-Exo4D adds synchronized views and expert commentary~\cite{grauman2024egoexo4d}; HoloAssist remote verbal guidance~\cite{wang2023holoassist}; and IndEgo co-located industrial collaboration~\cite{chavan2025indego}. Missing from these datasets is a co-located timeline linking task progress, requests, triggers, verbal or physical interventions, and outcomes. Without that linkage, a model may recognize an action or error but cannot learn how a helper's response depends on task state, initiation, and the assistance need.

DYAD provides that combination in 20 co-located gearbox-assembly sessions captured by a HoloLens~2 and Kinect workspace camera. One trained helper follows a guidance-first policy; mode labels therefore describe behavior under that policy, not a universal prescription. The corpus links 528 task-step intervals and 611 requests to 851 valid assistance records with trigger, initiation, hard-stop, object, utterance, and outcome annotations. Its contribution is interaction structure and sensing coverage, not scale.

To connect the corpus to the assistance loop, three benchmarks isolate distinct decisions: estimate current task state from causal video; select the demonstrated mode conditional on an upcoming intervention; and formulate a verbal response to a performer question. Evaluating them separately prevents task-perception, response-selection, and language-generation errors from being conflated; they are selected components, not an end-to-end assistant. Continuous need detection is not benchmarked because non-intervention negatives were not sampled; outcome prediction is excluded because 812 of 827 labeled outcomes are effective. Other annotations support future request grounding, trigger recognition, intervention timing, and embodiment-aware action selection.

This paper makes three contributions:

\begin{itemize}
    \item \textbf{Interaction-rich data and annotations.} Synchronized egocentric and workspace sensing connects task steps, requests, interventions, and outcomes; each intervention exposes mode, trigger, initiation, hard stops, objects, utterances, and outcome under one documented helper policy.
    \item \textbf{Component-level protocols.} Three leakage-aware tasks specify distinct questions, inputs, targets, participant splits, metrics, baselines, and limitations.
    \item \textbf{Diagnostic findings.} Step and response baselines remain limited under causal or privileged inputs; in mode anticipation, causal metadata (0.624 macro-F1) exceeds the strongest four-seed RGB mean (0.548), while a privileged trigger map reaches 0.915.
\end{itemize}

\section{Related Work}
\label{sec:related_work}

\paragraph{Procedural data}
Procedural datasets provide execution supervision at scale: Ego4D records long-form egocentric activity~\cite{grauman2022ego4d}; IKEA ASM and Assembly101 add multi-view assembly, action, and pose labels~\cite{ben2021ikea,sener2022assembly101}; IndustReal adds step-completion and error labels~\cite{schoonbeek2024industreal}; and Ego-Exo4D pairs first- and third-person views with expert commentary~\cite{grauman2024egoexo4d}. They describe what happened, but not whether help was requested, which modality a designated helper chose, or whether the response resolved the need.

\paragraph{Interactive assistance}
HoloAssist aligns actions, mistakes, dialogue, and intent and evaluates mistake detection, intervention type, and hand forecasting, but its instructor is remote and verbal-only~\cite{wang2023holoassist}. IndEgo records co-located peers and multimodal sensing without a fixed performer--helper policy or linked request--intervention--outcome events~\cite{chavan2025indego}. Vinci answers questions from live egocentric video rather than recording human modality choice~\cite{huang2025vinci}. DYAD adds a fixed helper role, linked events, and physical help; Table~\ref{tab:dataset-comparison} isolates this combination.
Vid2Coach transforms how-to videos into accessible task assistants and studies real-time remote human guidance~\cite{huh2025vid2coach}; DYAD centers on linked co-located requests, interventions, and outcomes.

\begin{table*}[t]
\centering
\scriptsize
\caption{Dataset comparison. Dashes denote the absence of the specified annotation structure, not necessarily the behavior. Outcomes refer to assistance-response outcomes, not task completion. Errors, mistakes, and intent are related labels rather than equivalent assistance triggers. Sensing includes captured streams and derived tracking or geometry.}
\label{tab:dataset-comparison}
\setlength{\tabcolsep}{3.2pt}
\begin{tabular}{@{}lccccccp{0.22\textwidth}@{}}
\toprule
Dataset & Co-located helper & Verbal & Physical & Requests & Triggers & Outcomes & Sensing \\
\midrule
Assembly101~\cite{sener2022assembly101} & -- & -- & -- & -- & -- & -- & Multi-view video, 3D hands \\
IndustReal~\cite{schoonbeek2024industreal} & -- & -- & -- & -- & Errors & -- & Ego RGB-D, stereo, gaze, hands, head pose \\
Ego-Exo4D~\cite{grauman2024egoexo4d} & -- & Commentary & -- & -- & -- & -- & Ego/exo RGB, gaze, 3D \\
HoloAssist~\cite{wang2023holoassist} & Remote & Yes & -- & Dialogue & Intent & -- & Ego RGB-D, gaze, hands \\
IndEgo~\cite{chavan2025indego} & Peers & Speech & Joint work & -- & Mistakes & -- & Ego/exo video, gaze, hand pose, 3D point cloud \\
\textbf{DYAD} & \textbf{Yes} & \textbf{Yes} & \textbf{Yes} & \textbf{Events} & \textbf{9 labels} & \textbf{Yes} & \textbf{Ego/exo RGB-D, gaze, hands, head pose} \\
\bottomrule
\end{tabular}
\end{table*}

\paragraph{Causal understanding and response evaluation}
Online recognition restricts inference to observations available at decision time~\cite{hu2022online}; observing an intervention would leak DYAD's mode target. Progress-aware online segmentation exposes uncertainty and over-segmentation in procedural steps~\cite{shen2024progress}, motivating causal inputs and both frame- and segment-level metrics.

PACE uses action-completion estimates to time proactive robotic assistance~\cite{de2025pace}. DYAD records procedural work with both cognitive and physical assistance needs. Mode prediction conditions on an upcoming event; need detection remains separate and requires sampling non-intervention negatives.

Reference similarity alone is incomplete for open responses. We combine contextual-embedding similarity from BERTScore~\cite{zhang2020bertscore} with a task-specific judge~\cite{liu2023geval,zheng2023llmjudge}; oracle context isolates response content from perception and dialogue-history errors.

\section{Dataset}
\label{sec:dataset}

\subsection{Collection Rationale and Protocol}
Human-human assistance exposes decisions absent from recordings of a person working alone: a helper observes progress and help seeking, decides whether and how to intervene, and grounds a response in the shared workspace. DYAD records these interactions as demonstrations and supervision for robotic and AR assistance research. Human physical actions are not robot commands; feasibility, safety, embodiment, and user preference remain separate requirements.

\begin{figure*}[t]
    \centering
    \includegraphics[width=0.78\textwidth]{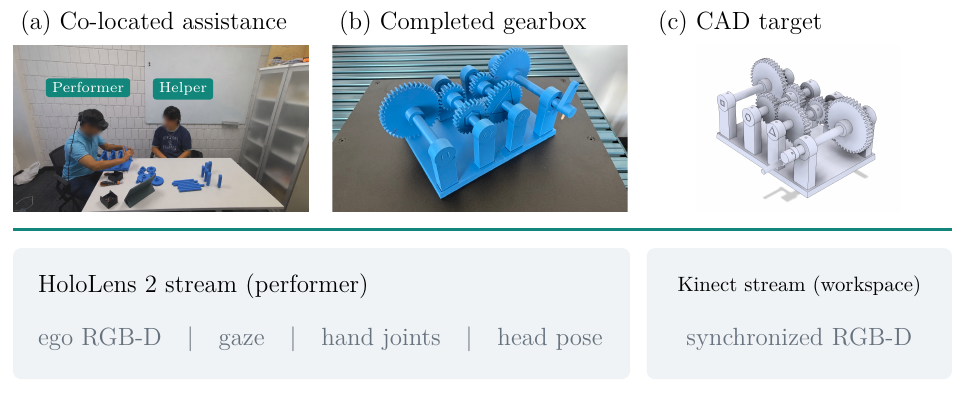}
    \caption{Collection setup and task apparatus. (a) Kinect view of the co-located performer--helper interaction; faces are blurred for privacy. (b) Completed physical gearbox. (c) CAD target shown to performers. Bottom: synchronized release modalities.}
    \label{fig:gearbox}
\end{figure*}

Each session pairs a HoloLens~2 wearer (the \emph{performer}) assembling a physical gearbox from a reference CAD model (Fig.~\ref{fig:gearbox}) with a nearby \emph{helper}. Similar parts, partial ordering constraints, out-of-reach items, and unstable subassemblies elicit search, confusion, ordering, reach, and stabilization needs without scripting individual mistakes. Six pilot sessions were used to refine the task and helper protocol before the 20 recorded sessions, which total 5\,h 58\,min (mean 17\,min 55\,s; range 8\,min 57\,s--29\,min 40\,s).

\paragraph{Single-helper policy}
The same trained helper conducts all sessions under one guidance-first policy: prefer verbal guidance, allow 5--10\,s for self-correction, intervene after a visible stall, use physical help for reach or stabilization, and avoid repeated cues absent an immediate error. Mode labels therefore describe demonstrated behavior under this policy, not a universal prescription.

\subsection{Sensing, Release, and Annotations}
Streams are synchronized through Platform for Situated Intelligence~\cite{bohus2021platform}. Table~\ref{tab:modalities} reports capture properties. The research release will provide deidentified ego RGB and depth, gaze, hand joints, head pose, Kinect RGB-D, timestamps, annotations, manifests, task materials, and benchmark code under controlled access. Audio was collected but will not be released; deidentified request and helper transcripts will be provided. Faces are blurred. Participants gave informed consent under an approved institutional protocol that permits deidentified research sharing.

\begin{table}[t]
\centering
\footnotesize
\caption{Synchronized modalities included in the planned release.}
\label{tab:modalities}
\resizebox{\columnwidth}{!}{%
\begin{tabular}{@{}lll@{}}
\toprule
Stream & Rate & Representation \\
\midrule
Ego RGB & 24--30\,Hz & $896\times504$ frames \\
Ego depth & 5\,Hz & $320\times288$, 16-bit \\
Gaze & 30\,Hz & 3D origin/direction \\
Hands & 20\,Hz & 26 joints/hand, $4\times4$ poses \\
Head pose & 30\,Hz & $4\times4$ transform \\
Kinect RGB & 15\,Hz & $1280\times720$ JPEG \\
Kinect depth & 15\,Hz & Native depth frames \\
\bottomrule
\end{tabular}%
}
\end{table}

Three aligned annotation layers describe \textbf{task steps} (528 intervals over eight coarse steps), \textbf{performer requests} (611 transcribed intervals), and \textbf{assistance events} (851 valid interventions). The assistance-event schema includes onset/offset, involved objects, helper utterance, and: \emph{mode} (verbal guidance or physical help); \emph{initiation} (performer-requested or helper-initiated); \emph{hard stop} (an explicit corrective pause); \emph{outcome} (effective, ineffective, or unnecessary); and one of nine triggers: out of reach, cannot find part, cannot decide next step, wrong part, ordering violation, wrong orientation, wrong screwdriver, stabilization needed, or missed screw.

\begin{table}[t]
\centering
\footnotesize
\caption{Compact annotation schema. Nullable fields preserve missing or out-of-taxonomy values rather than being imputed.}
\label{tab:schema}
\begin{tabular}{@{}p{0.16\columnwidth}p{0.30\columnwidth}p{0.40\columnwidth}@{}}
\toprule
Layer & Principal fields & Operational unit \\
\midrule
Task step & step, objects, tool & Current coarse assembly interval \\
Request & performer utterance, confidence & Explicit question or request interval \\
Assistance & mode, initiation, hard stop, trigger, objects, helper utterance, outcome & Delivered helper intervention \\
\bottomrule
\end{tabular}
\end{table}

\begin{table}[t]
\centering
\footnotesize
\caption{Closed eight-label task-step vocabulary. Counts are annotated intervals; one of 528 intervals has a null step label.}
\label{tab:steps}
\begin{tabular}{@{}lrlr@{}}
\toprule
Step label & $N$ & Step label & $N$ \\
\midrule
Gear on shaft & 108 & Part search & 83 \\
Bearing in holder & 96 & Shaft assembly in post & 28 \\
Second holder & 89 & Attach crank & 20 \\
First holder & 87 & Verify rotation & 16 \\
\bottomrule
\end{tabular}
\end{table}

\begin{table*}[t]
\centering
\footnotesize
\caption{Assistance-trigger vocabulary. A null trigger denotes a need outside this nine-label taxonomy, not a negative event.}
\label{tab:triggers}
\begin{tabular}{@{}p{0.145\textwidth}rp{0.27\textwidth}p{0.145\textwidth}rp{0.27\textwidth}@{}}
\toprule
Trigger & $N$ & Operational meaning & Trigger & $N$ & Operational meaning \\
\midrule
Out of reach & 172 & Required part/tool is beyond comfortable reach. &
Cannot find part & 124 & Required part cannot be located or distinguished. \\
Cannot decide next step & 123 & Performer is visibly uncertain about the next action. &
Wrong part & 96 & An incorrect component is selected or used. \\
Ordering violation & 63 & Attempt conflicts with a required dependency or order. &
Wrong orientation & 63 & Component pose or orientation is incorrect. \\
Wrong screwdriver & 57 & Tool or driver head is inappropriate for the fastener. &
Stabilization needed & 36 & A second person must hold a part or subassembly. \\
Missed screw & 22 & A required fastening action was omitted. & & & \\
\bottomrule
\end{tabular}
\end{table*}

\paragraph{Annotation and cleaning}
Annotators labeled all layers in PSI Studio while viewing synchronized streams. Requests and interventions remain separate because the helper can act without an explicit request and one request can prompt multiple actions. The release pipeline converts every session to a relative clock, preserves null timestamps, normalizes controlled-vocabulary spelling, checks nonnegative durations and legal values, and generates participant-disjoint manifests. Two source sessions used absolute-style hour fields but retained valid within-session timing; their session-specific clock origins are stored rather than treated as elapsed hours.

\paragraph{Reliability}
A second annotator independently and blindly labeled one complete session and prespecified continuous windows from seven additional sessions. Timestamps were aligned separately for each session, assistance events were matched one-to-one within 1\,s, and agreement was computed before adjudication using detection measures, boundary errors, and Cohen's $\kappa$~\cite{cohen1960coefficient} (Table~\ref{tab:irr}). The audit covered eight participants and 147 reference assistance events (17.3\% of the corpus), of which 144 were matched. The added windows also audited the task-step and request streams.

\begin{table}[t]
\centering
\footnotesize
\caption{Independent, pre-adjudication reliability audit. Null is a substantive trigger category.}
\label{tab:irr}
\begin{tabular}{@{}p{0.62\columnwidth}r@{}}
\toprule
Measure & Result \\
\midrule
Sessions / participants & 8 / 8 \\
Reference / matched assistance events & 147 / 144 \\
Assistance-event corpus coverage & 17.3\% \\
Assistance detection P / R / F1 & 1.000 / 0.980 / 0.990 \\
Mode agreement / $\kappa$ & 100\% / 1.000 \\
Hard-stop agreement / $\kappa$ & 100\% / 1.000 \\
Trigger agreement / $\kappa$ & 84.6\% / 0.826 \\
Added-window step name / tool $\kappa$ ($n=32$) & 1.000 / 0.908 \\
Added-window request matches; start / end MAE & 50/50; 0.047 / 0.206\,s \\
\bottomrule
\end{tabular}
\end{table}

\begin{table}[t]
\centering
\footnotesize
\caption{Corpus statistics and benchmark eligibility.}
\label{tab:stats}
\begin{tabular}{@{}lr@{}}
\toprule
Sessions / duration & 20 / 5\,h 58\,min \\
Steps / requests & 528 / 611 \\
Valid assistance records & 851 \\
Mode-label availability & 850 / 851 \\
Verbal / physical & 619 / 231 \\
Initiation-label availability & 850 / 851 \\
Performer- / helper-initiated & 580 / 270 \\
Hard stops & 135 \\
Outcome-label availability & 827 / 851 \\
Effective / ineffective / unnecessary & 812 / 3 / 12 \\
\midrule
Complete 3\,s RGB + 0.5\,s gap & 829 (610 / 219) \\
Step track & 20 sessions \\
Response track & 376 pairs \\
\bottomrule
\end{tabular}
\end{table}

\paragraph{Observed interaction structure}
Verbal guidance is 2.7 times as common as physical help; 580 of 850 labeled events are performer-initiated, and 756 have triggers. Reach and stabilization are about 92\% physical, whereas decision, ordering, and orientation problems are 95--100\% verbal. Out-of-reach help is usually requested; ordering violations are commonly caught by the helper. Because 812 of 827 labeled outcomes are effective, we do not benchmark outcome prediction.

\paragraph{Count reconciliation}
The export contains 852 rows. Removing one malformed duplicate leaves 851 valid events. Mode analysis uses the 850 records with mode labels; 21 lack a complete synchronized 3\,s pre-onset RGB window with a 0.5\,s gap, leaving 829 mode examples. This includes five terminal events after the final recorded RGB frame. All 20 performers remain in the participant-grouped mode folds, and all 20 sessions enter the step track. The response track retains requests matched to verbal replies.

\section{Reference Tasks}
\label{sec:benchmark-suite}

The task suite follows the assistance loop defined in Section~\ref{sec:introduction}. It evaluates three components supported by the annotations: task-state estimation, assistance-mode selection conditional on an upcoming intervention, and verbal response generation. Assistance-need detection is not benchmarked because the present protocols do not sample non-intervention intervals as negatives. Table~\ref{tab:benchmark-suite} makes each task's question, input, split, leakage boundary, metric, and limitation explicit. The tasks are not stages of a measured end-to-end system and have no shared score.

\begin{table*}[t]
\centering
\scriptsize
\caption{Reference-task contracts. Oracle denotes annotated task and interaction context unavailable to a deployable system.}
\label{tab:benchmark-suite}
\setlength{\tabcolsep}{3pt}
\begin{tabular}{@{}p{0.12\textwidth}p{0.28\textwidth}p{0.27\textwidth}p{0.26\textwidth}@{}}
\toprule
Task & Question and input $\rightarrow$ target & Cohort, split, and leakage boundary & Metrics and limitation \\
\midrule
Step understanding & What is the performer doing? Causal ego RGB $\rightarrow$ 8 step labels. & 5 validation / 15 test sessions; future frames and assistance labels withheld. & Segment F1, edit, localization mAP, frame metrics. One procedure only. \\
Mode anticipation & Given an upcoming event, speak or act? Pre-onset ego RGB $\rightarrow$ demonstrated mode. & 20 people; grouped 5-fold CV; $N{=}829$; current-event labels and speech withheld. & Macro-F1 and physical AUPRC. Does not detect need or judge appropriateness. \\
Response generation & What should the helper say? Question + oracle context $\rightarrow$ utterance. & 5 development / 15 test sessions; 120 / 256 pairs; current trigger withheld. & Judge rubric and BERTScore. Context is privileged. \\
\bottomrule
\end{tabular}
\end{table*}

These contracts prevent post-hoc trigger labels or ground-truth dialogue history from being mistaken for causal inputs. They also identify the mode target as one helper's observed action under the collection policy, not a universal prescription.

\subsection{Assistance-Mode Anticipation}
\label{sec:benchmark}

\paragraph{Question and target}
For an assistance event with onset $t_i$, the model observes
\[
X_i^{w,g}=X[t_i-g-w,\;t_i-g]
\]
and predicts the helper's demonstrated mode $y_i\in\{\texttt{verbal},\texttt{physical}\}$. We use a $w{=}3$\,s window ending $g{=}0.5$\,s before onset. The gap reduces leakage from the helper's first reach or utterance. The task assumes an intervention will occur; it does not detect assistance need.

\paragraph{Input, split, and leakage boundary}
The standard input is egocentric RGB; synchronized hands and gaze are optional ablations. Current-event trigger, initiation, outcome, speech, task-step label, participant identity, and capture filename are prohibited. All methods use the same cleaned 829-event manifest and participant-grouped five-fold split, with 12/4/4 train/validation/test performers per fold and no identity overlap.

\paragraph{Metrics and baselines}
Macro-F1 is primary because physical assistance is the minority class; physical-class AUPRC is secondary, with per-class F1 and balanced accuracy retained as diagnostics. Participant bootstrap intervals account for repeated events within a session. Majority prediction is fitted on each training fold. ResNet18-last~\cite{he2016deep} encodes the final admissible frame, ResNet18-pool averages eight frame features, R3D-18~\cite{tran2018closer} and X3D-S~\cite{feichtenhofer2020x3d} are 3D CNNs, and VideoMAE~\cite{tong2022videomae} encodes the clip. Learned visual models use class-weighted cross-entropy and participant-disjoint model selection; RGB results average seeds $\{0,1,2,42\}$. Frozen RGB features are also fused with hand/gaze encoders or a six-dimensional causal metadata vector containing clock time, session progress, event index, and prior-assistance counts.

We include two controls with different information contracts. A causal metadata logistic receives timing and prior counts but no pixels, current trigger, step label, or transcript. A privileged trigger mapping predicts physical for reach and stabilization triggers and verbal otherwise. Because triggers were assigned after annotators saw the completed intervention, this mapping measures annotation consistency rather than deployable prediction.

\paragraph{Reproducibility protocol}
Released manifests identify each admissible clip and rotate four-person test and validation groups, placing every performer in one out-of-fold test set. Validation participants select models. Stored predictions retain labels, event IDs, and seeds; scores are recomputed over 829 unique out-of-fold events, making eligibility, duplicate removal, and participant bootstraps auditable.

\begin{table}[t]
\centering
\footnotesize
\caption{Mode anticipation ($N{=}829$). RGB: four-seed mean$\pm$sample std; fusion: seed 42.}
\label{tab:main-results}
\begin{tabular}{@{}lcc@{}}
\toprule
Method & Macro-F1 & Phys. AUPRC \\
\midrule
Majority & 0.424 & 0.264 \\
ResNet18-last & $0.534{\pm}0.014$ & $0.311{\pm}0.018$ \\
ResNet18-pool & $0.529{\pm}0.015$ & $0.310{\pm}0.011$ \\
R3D-18 & $0.522{\pm}0.019$ & $0.286{\pm}0.020$ \\
X3D-S & $0.548{\pm}0.007$ & $0.377{\pm}0.012$ \\
VideoMAE & $0.545{\pm}0.017$ & $0.315{\pm}0.009$ \\
\midrule
VideoMAE + hands/gaze & 0.544 & 0.327 \\
Metadata & 0.624 & 0.442 \\
Metadata + RGB scores & 0.625 & 0.446 \\
\midrule
Trigger oracle (privileged) & 0.915 & 0.809 \\
\bottomrule
\end{tabular}
\end{table}

\paragraph{Multimodal and diagnostic analyses}
The synchronized streams permit tests beyond RGB-only ranking. For each of two frozen visual backbones, Table~\ref{tab:multimodal} compares RGB with hands, gaze, and both auxiliary streams under the identical split and seed. Comparisons are made within this fusion trainer rather than against the separately trained four-seed RGB rows. We also compare the causal metadata and seed-42 VideoMAE models within the three most physically informative populated triggers (Table~\ref{tab:trigger-gap}). This analysis asks whether an aggregate difference is concentrated in a particular assistance need rather than implying that post-hoc triggers are available at inference.

\begin{table}[t]
\centering
\footnotesize
\caption{Frozen-RGB late fusion (seed 42); $\Delta$ uses the same trainer's RGB-only run.}
\label{tab:multimodal}
\begin{tabular}{@{}llcc@{}}
\toprule
Backbone & Input & Macro-F1 & $\Delta$ \\
\midrule
ResNet-pool & RGB & 0.527 & -- \\
 & RGB + hands & 0.483 & $-0.045$ \\
 & RGB + gaze & 0.510 & $-0.018$ \\
 & RGB + hands + gaze & 0.533 & $+0.006$ \\
\midrule
VideoMAE & RGB & 0.532 & -- \\
 & RGB + hands & 0.531 & $-0.002$ \\
 & RGB + gaze & 0.528 & $-0.005$ \\
 & RGB + hands + gaze & 0.544 & $+0.011$ \\
\bottomrule
\end{tabular}
\end{table}

\begin{table}[t]
\centering
\footnotesize
\caption{Trigger diagnostic ($N{=}829$); triggers are not model inputs.}
\label{tab:trigger-gap}
\resizebox{\columnwidth}{!}{%
\begin{tabular}{@{}lrrrrr@{}}
\toprule
Trigger & $n$ & Meta rec. & RGB rec. & Meta F1 & RGB F1 \\
\midrule
\texttt{out\_of\_reach} & 160 & .466 & .426 & .624 & .589 \\
\texttt{stabilization\_needed} & 36 & .515 & .333 & .667 & .468 \\
\texttt{cannot\_find\_part} & 123 & .793 & .103 & .561 & .109 \\
\bottomrule
\end{tabular}%
}
\end{table}

\paragraph{Results and limitations}
Across seeds, X3D-S has the highest mean RGB macro-F1 ($0.548{\pm}0.007$); its seed-42 margin over majority is 0.122 (participant-bootstrap 95\% CI $[0.054,0.196]$). Causal metadata reaches 0.624 and exceeds seed-42 VideoMAE by 0.095 $[0.027,0.157]$. RGB score stacking reaches 0.625 but does not improve reliably over metadata alone ($-0.001$ $[-0.054,0.049]$). Hand/gaze fusion is mixed: the strongest cell reaches 0.544, $+0.011$ over its same-trainer RGB run. The privileged trigger map reaches 0.915 but is an annotation upper bound. For \texttt{cannot\_find\_part}, VideoMAE physical recall is 0.103 versus metadata's 0.793.

Scores aggregate 829 unique OOF predictions; intervals use 2,000 participant-level replicates. The target remains one helper's action under one policy, not universal appropriateness or robot feasibility. Because every example is an assistance event, the task cannot establish whether or when help is needed.

\subsection{Causal Assembly-Step Understanding}
\label{sec:companion:steps}

\paragraph{Protocol}
The question is: What is the performer doing now? Input is ego RGB sampled at 1\,Hz; the target is one of eight assembly steps at each decision time. Predictions are aggregated into temporal segments describing when each step begins and ends. Only frames at or before the decision time are permitted; future frames, requests, assistance labels, and helper utterances are withheld. Thus the task measures both causal state recognition and boundary estimation rather than offline video classification. Five of the 20 sessions form the validation split and 15 form the participant-disjoint test split. Segmental F1@0.1, sequence edit, temporal mAP@0.1, frame mAP, and mean-over-frames (MoF) capture complementary errors.

\paragraph{Baselines and results}
Frozen SmolVLM2~\cite{marafioti2025smolvlm} and Qwen3-VL~\cite{bai2025qwen3vl} models score a causal four-frame lookback against short captions for the fixed vocabulary. Pointwise mutual-information debiasing subtracts each caption's score on blank gray frames, reducing preference for inherently likely wording. A causal decoder then applies exponential smoothing, hysteresis, and minimum dwell: a competing step must exceed the current step by a margin and remain dominant before a transition is committed. Table~\ref{tab:eval-steps} shows that no model dominates every metric. Qwen3-VL-2B gives the highest segment F1 (17.4), Qwen3-VL-8B the highest edit score (28.3), and Qwen3-VL-4B the highest localization mAP (8.9). The majority baseline's stronger frame metrics reflect class prevalence rather than useful temporal localization.

MoF measures frame accuracy and frame mAP per-step ranking. Segmental F1@0.1 matches intervals at 0.1 IoU, edit measures order after collapsing repeats, and temporal mAP@0.1 evaluates localization. Both families matter because frequent-step predictions can score well framewise while missing boundaries and order.

\begin{table}[t]
\centering
\footnotesize
\caption{Zero-shot causal step understanding on 15 held-out sessions.}
\label{tab:eval-steps}
\resizebox{\columnwidth}{!}{%
\begin{tabular}{@{}lrrrrr@{}}
\toprule
& F1@0.1 & Edit & Loc. mAP & Frame mAP & MoF \\
\midrule
Majority & 1.9 & 4.3 & 0.4 & \textbf{24.8} & \textbf{26.5} \\
SmolVLM2-500M & 10.3 & 23.4 & 4.4 & 19.9 & 8.1 \\
SmolVLM2-2.2B & 11.4 & 27.8 & 3.9 & 19.5 & 10.5 \\
Qwen3-VL-2B & \textbf{17.4} & 26.1 & 7.5 & 20.6 & 22.1 \\
Qwen3-VL-4B & 14.7 & 22.6 & \textbf{8.9} & 20.1 & 15.4 \\
Qwen3-VL-8B & 15.8 & \textbf{28.3} & 7.2 & 19.6 & 16.1 \\
\bottomrule
\end{tabular}%
}
\end{table}

\begin{table}[t]
\centering
\footnotesize
\caption{Qwen3-VL-2B causal-decoder ablation.}
\label{tab:step-ablation}
\resizebox{\columnwidth}{!}{%
\begin{tabular}{@{}lrrrrr@{}}
\toprule
Variant & F1@0.1 & Edit & Loc. mAP & Frame mAP & MoF \\
\midrule
Full decoder & 17.4 & 26.1 & 7.5 & 20.6 & 22.1 \\
Without EMA & 16.1 & 24.1 & 7.2 & 18.5 & 21.8 \\
Without dwell & 15.6 & 20.2 & 8.2 & 20.6 & 22.5 \\
Without PMI & 12.4 & 19.8 & 5.8 & 21.0 & 19.0 \\
\bottomrule
\end{tabular}%
}
\end{table}

Removing PMI reduces segment F1 from 17.4 to 12.4, and removing minimum dwell reduces it to 15.6 while slightly increasing localization mAP and MoF. Removing EMA lowers every reported metric. The ablation reinforces why temporal and frame metrics must be reported together.

\paragraph{Limitation}
The closed vocabulary and single procedure do not test transfer, assistance-need detection, or intervention quality.

\subsection{Instructor Response Generation}
\label{sec:companion:qa}

\begin{table}[t]
\centering
\scriptsize
\caption{Response-generation results. Pair-level means; judge metrics use a 0--5 scale.}
\label{tab:companion-qa}
\resizebox{\columnwidth}{!}{%
\begin{tabular}{@{}llrrrrrr@{}}
\toprule
Split & Model & Corr. & Comp. & Rel. & Clear & Brief & Final \\
\midrule
Dev. ($N{=}120$) & GPT-4o & 3.750 & 3.600 & 3.908 & 4.300 & 4.925 & 3.567 \\
 & GPT-4.1 & 4.050 & 3.925 & 4.142 & 4.533 & 4.525 & 3.808 \\
\midrule
Test ($N{=}256$) & GPT-4o & 3.836 & 3.629 & 3.953 & 4.270 & 4.914 & 3.582 \\
 & GPT-4.1 & 4.031 & 3.918 & 4.094 & 4.500 & 4.512 & 3.688 \\
\bottomrule
\end{tabular}%
}
\vspace{1mm}

\resizebox{0.72\columnwidth}{!}{%
\begin{tabular}{@{}llrrr@{}}
\toprule
Split & Model & BERT-P & BERT-R & BERT-F1 \\
\midrule
Dev. ($N{=}120$) & GPT-4o & 0.889 & 0.862 & 0.875 \\
 & GPT-4.1 & 0.855 & 0.857 & 0.856 \\
\midrule
Test ($N{=}256$) & GPT-4o & 0.888 & 0.862 & 0.874 \\
 & GPT-4.1 & 0.861 & 0.857 & 0.858 \\
\bottomrule
\end{tabular}%
}
\end{table}

\paragraph{Protocol}
Given a performer's question, the model generates a concise, contextually grounded helper utterance. This benchmark evaluates response generation when verbal assistance is appropriate; it does not decide whether assistance should be verbal or physical. Requests are matched to the temporally closest assistance event from 5\,s before to 15\,s after request end, accommodating annotation offsets and natural response delay. Only matches containing a verbal helper utterance are retained; requests answered solely through physical action are excluded rather than assigned an artificial spoken target. Five development sessions contain 120 pairs and 15 participant-disjoint test sessions contain 256.

The structured oracle context provides: the procedure, part names, and common errors; all annotated task steps up to the question with completion status and objects; ten previous dialogue turns using ground-truth helper responses; ten previous physical interventions; a gearbox reference image; and the current question. The current trigger is withheld because it was annotated after the response, although trigger labels may occur in prior-intervention history. Ground-truth task and dialogue history isolate response quality from upstream perception and from compounding a model's own earlier errors. This is deliberately not a deployable video-to-response setting.

\paragraph{Baselines, metrics, and results}
GPT-4o~\cite{openai2024gpt4o} and GPT-4.1~\cite{openai2025gpt41} use identical prompts, temperature 0, one generation per query, and a 100-token cap; they are instructed to answer as a real-time helper in one or two actionable sentences. Claude Opus 4.5~\cite{anthropic2025opus45} (temperature 0), from a different provider and model family, receives the same oracle context and human response as reference. We adapt the context-aware evaluation approach of Qorbani et al.~\cite{qorbani2025contextaware}: the judge scores correctness, completeness, contextual relevance, and actionable clarity, with added brevity and holistic final scores, each from 0--5. BERTScore precision, recall, and F1 provide complementary reference similarity.

The five development sessions were used for prompt engineering; the held-out sessions were not used to revise the prompt or evaluation. Table~\ref{tab:companion-qa} reports pair-level means. GPT-4.1 has the higher task-aware final score, while GPT-4o is briefer and has higher contextual-embedding similarity to the reference; we treat them as representative baselines, not a model ranking. Final-score means vary substantially by session (GPT-4o: 2.756--4.364; GPT-4.1: 2.833--4.364), indicating interaction-dependent difficulty. The difference between task-aware scores and reference similarity also reflects that an appropriate reply need not reproduce the helper's wording.

\paragraph{Limitation}
The task excludes requests answered only through physical action, tests two proprietary model families, and uses an LLM judge not directly validated by human ratings on DYAD. Participant-level uncertainty and evaluation with observable rather than annotated context remain open extensions.

\FloatBarrier

\section{Discussion}
\label{sec:discussion}

\paragraph{Why use DYAD?}
Assembly101 offers scale and action diversity, Ego-Exo4D broader activities and viewpoints, HoloAssist remote verbal instruction, and IndEgo co-located peer collaboration. DYAD instead centers the assistance interaction, linking a designated helper's verbal and physical interventions to requests, task state, triggers, and outcomes. This supports need detection, request grounding, intervention timing, and mode selection without reconstructing interactions across corpora.

Reference tasks use separate slices---causal ego video and steps; pre-onset sensing and modes; or questions and structured history---and no benchmark consumes every field. The release also supports continuous need detection, nine-way trigger recognition, request--response matching, hard-stop prediction, object grounding, and sensing ablations once new evaluation protocols are defined.

\paragraph{What the reference tasks show}
The step results show weak temporal localization despite a majority baseline with higher frame accuracy. Response generation remains imperfect even with oracle task and interaction context. For mode anticipation, the ordering is privileged trigger map $\gg$ causal metadata $>$ RGB; RGB stacking does not reliably improve on metadata. This indicates recoverable session structure, not that vision is irrelevant. Weak VideoMAE physical recall for within-search events further shows that pre-onset appearance alone does not explain the helper's policy. The three tasks have different contracts and are not stages of an evaluated end-to-end assistant.

\paragraph{From mode prediction to assistive policy}
A deployable assistant could use mode prediction as a gate: verbal decisions route to dialogue or AR guidance, while physical decisions route to action selection. Combined with need and timing models such as PACE~\cite{de2025pace}, this could support context-sensitive prompts, proactive handovers, and stabilization. The latter two require different branches---grasp-and-handover versus hold/brace---with distinct grounding, control, and safety assumptions. DYAD labels can supervise such routing, but not motor-skill execution; need inference, action planning, and safety validation remain unbenchmarked, and human actions are not robot commands.

\paragraph{Interaction patterns and collection lessons}
Assistance events outnumber explicit requests by 1.39:1, and 270 of 851 interventions are helper-initiated. Initiation is structured by need: out-of-reach help is usually requested, whereas ordering violations are often caught proactively. Hard stops likewise concentrate in ordering and orientation errors rather than reach or part search. These policy-specific associations show the value of linking requests, triggers, initiation, modes, and outcomes.

The annotation export lacked persistent event IDs and mixed session-relative with absolute-style clocks, requiring ID reconstruction and stored clock transforms before windowing. Future collections should record both at capture time, distinguish missing from out-of-taxonomy values, and permit nested or multilabel assistance when speech and physical help overlap.

\paragraph{Limitations}
DYAD contains 20 sessions from one gearbox task, site, and helper. Participant-grouped evaluation tests new performers within that setting, not transfer to new tasks, workspaces, populations, helpers, or embodiments. The single-helper design controls policy variation but also entangles mode labels with that person's habits. Physical assistance is the minority class, and several triggers have limited support. The reliability audit spans eight sessions from one primary annotator's subset and therefore measures agreement for this annotator pair.

Mode anticipation conditions on a known intervention and cannot detect need; step understanding assumes a fixed vocabulary; and response generation uses privileged context and an LLM judge not directly validated on DYAD. Outcome prediction is not reported because 812 of 827 labeled outcomes are effective; meaningful modeling requires targeted failure and recovery collection.

\section{Conclusion}
\label{sec:conclusion}

DYAD records 20 performers assisted by one co-located human helper, linking 528 task steps and 611 requests to 851 assistance events under synchronized sensing. Three reference tasks target selected components: causal step understanding reaches 17.4 segment F1@0.1, instructor-response baselines 3.58--3.69/5 under oracle context, and mode anticipation on 829 eligible events $0.548{\pm}0.007$ macro-F1 with RGB and 0.624 with causal metadata, below a 0.915 privileged trigger upper bound. The timeline supports future benchmarks for need detection, grounding, timing, triggers, and embodied assistance; outcome prediction requires more diversity. DYAD makes these components observable before integration into robotic or AR assistance policies.

\bibliographystyle{IEEEtran}
\bibliography{bib/references}

\end{document}